\documentclass[times,art10,twocolumn,latex8]{article}

\usepackage{iccv}
\usepackage{times}
\usepackage{epsfig}
\usepackage{graphicx}
\usepackage{amsmath}
\usepackage{amssymb}
\usepackage{flushend}

\usepackage{breqn}
\usepackage{float}
\usepackage{balance}

\usepackage[pagebackref=true,breaklinks=true,letterpaper=true,colorlinks,bookmarks=false]{hyperref}

\iccvfinalcopy 

\begin{document}

\title{Robust pose tracking with a joint model of appearance and shape}

\author{Yuliang Guo \qquad Lakshmi Narasimhan Govindarajan \qquad Benjamin Kimia \qquad Thomas Serre \\
{\tt\small \{yuliang_guo,lakshmi_govindarajan,benjamin_kimia,thomas_serre\}@brown.edu} \\
Brown University, Providence, RI, USA\\
}
\maketitle

\begin{abstract}
We present a novel approach for estimating the 2D pose of an articulated object with an application to automated video analysis of small laboratory animals. We have found that deformable part models developed for humans -- exemplified by the flexible mixture of parts (FMP) model \cite{Yang_Ramanan_pami2013} -- typically fail on challenging animal poses. We argue that beyond encoding appearance and spatial relations, shape is needed to overcome the lack of distinctive landmarks on laboratory animal bodies. In our approach, a shape consistent FMP (scFMP) model computes promising pose candidates after a standard FMP model is used to rapidly discard false part detections. This ``cascaded'' approach combines the relative strengths of spatial-relations, appearance and shape representations and is shown to yield significant improvements over the original FMP model as well as a representative deep neural network baseline \cite{Wei2016-ov}. 
\end{abstract}

\section{Introduction}

The increasing use of animal models in biomedical research has led to new demands for high-throughput automated testing methodologies capable of assaying complex behaviors \cite{Schaefer2012-mx}. Accurately tracking an animal pose is key to assaying a number of behaviors for different animal models including orienting in larvae \cite{Gomez-Marin2012-aw}, swimming and other locomotory behaviors in nematodes \cite{Stephens2008-wv,Restif2014-cn}, and social behaviors and object recognition in rodents \cite{Patel2014-gw}.  

Several systems have been developed to robustly track large groups of small laboratory animals, see \eg \cite{Reiser2009-sg,Pelkowski2011-tb,Albrecht2011-ju,Ohayon2013-jh,Weissbrod2013-oh,Shen2015-py}. However, most of these systems are limited to tracking a center of mass and do not allow for a finer analysis of posture. Several methods have also been specifically developed for automatically estimating the pose of small laboratory animals, see \eg \cite{Gomez-Marin2012-aw,Patel2014-gw,Restif2014-cn,Jung2014-ou,Palmer2016-qe}. For the most part, these methods rely on simple image processing (\eg background subtraction) to extract the silhouette of a body before computing a medial axis transform. 

\begin{figure*}[t!]
\begin{center}
   \includegraphics[width=0.135\linewidth]{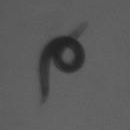}
   \includegraphics[width=0.135\linewidth]{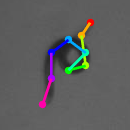}
   \includegraphics[width=0.135\linewidth]{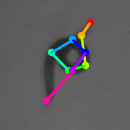}
   \includegraphics[width=0.135\linewidth]{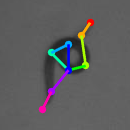}
   \includegraphics[width=0.135\linewidth]{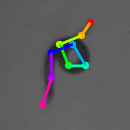}
   \includegraphics[width=0.135\linewidth]{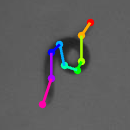}
   \includegraphics[width=0.135\linewidth]{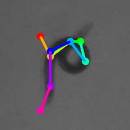}\\
   \includegraphics[width=0.135\linewidth]{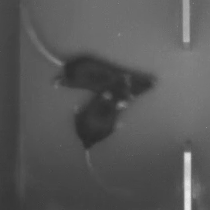}
   \includegraphics[width=0.135\linewidth]{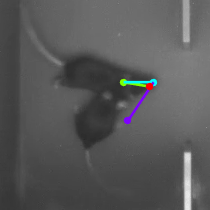}
   \includegraphics[width=0.135\linewidth]{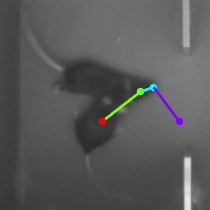}
   \includegraphics[width=0.135\linewidth]{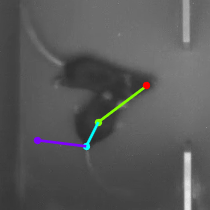}
   \includegraphics[width=0.135\linewidth]{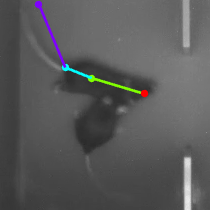}
   \includegraphics[width=0.135\linewidth]{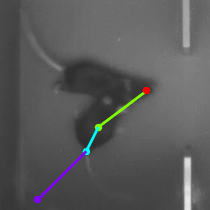}
   \includegraphics[width=0.135\linewidth]{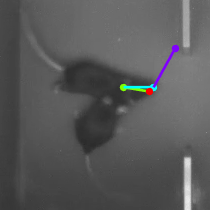}
\end{center}
   \caption{Representative FMP ranking errors of pose configurations for a nematode (during an omega turn; top row) and two socially-interacting rodents (bottom row). From left to right are rank-ordered part configurations for the same frame (leftmost column). False configurations are consistently ranked higher than the correct ones (fourth pose for the top row and entirely missed for the bottom row).}
\label{fig:MixtureOfPartsFail}
\end{figure*}

A major drawback of such methods is their inability to discriminate between the front and rear ends of the body, forcing researchers to rely on simple heuristics instead \cite{Restif2014-cn,Jung2014-ou} (\eg by computing the direction of movement and assuming that the animal moves forward). In addition, background subtraction tends to be sensitive to changes in illumination and often yields erroneous pose estimates. In the context of biomedical research, these failures need to be detected -- either automatically \cite{Restif2014-cn} or manually \cite{Stephens2008-wv,Jung2014-ou} in order to exclude the corresponding frames from further behavioral analysis. Such an opportunistic approach to pose estimation may lead to significant biases in behavioral analyses if those system failures tend to co-occur more frequently with certain behaviors (\eg for those behaviors that yield significant self-occlusion such as omega turns in nematodes or social interactions in rodents, see Fig.~\ref{fig:MixtureOfPartsFail}). Overall, existing approaches are not robust enough to allow for the throughput behavioral analysis needed for modern biomedical research applications. %
 
Learning-based approaches on the other hand have recently lead to significant improvements in human pose estimation (see \cite{Sarafianos2016-wo} for a review). Such approaches fall under two broad classes: structured probabilistic models and deep learning-based methods. A representative structured probabilistic approach is the Flexible Mixture of Parts (FMP) model \cite{Yang_Ramanan_pami2013}. This model extends earlier part-based approaches (\eg \cite{Andriluka2009-sx,Singh2010-ka}) -- jointly encoding the appearance of body parts together with their spatial relations. The FMP model uses rather simple body-part detectors (mixtures of HoG templates) but explicitly encodes the relative placement of adjacent parts to constrain the overall body configuration. 

Advances in deep convolution networks (DCNs) have allowed the training of more robust body-part detectors -- leading to approaches that, unlike earlier structured probabilistic models, rely exclusively on appearance to estimate pose -- bypassing the need for spatial constraints. Representative DCNs for the detection of a single person includes DeepPose \cite{Toshev2014-hs} and Convolutional Pose Machines (CPMs) \cite{wei2016cpm}. In essence, a convolutional pose machine is a sequence of deep networks that each produce confidence maps for individual parts. Every stage in this sequence receives image features as well as confidence maps from the previous stage as input, with partial supervision at each stage. Spatial constraints between parts are only implicit, image-dependent and learned from intermediate confidence maps. 

More recently, researchers have started to focus on the challenge presented by the simultaneous pose tracking of multiple people in videos. OpenPose \cite{Cao:etal:CVPR2017} was developed as a CPM extension specifically to improve the efficiency by which body parts get assigned to appropriate individuals. Here, since our challenging small laboratory animal dataset is comprised of only a single individual per frame, we elected to use CPM as a baseline. 

Adapting current methods developed for humans to animal pose estimation brings several challenges. In contrast to human bodies for which multiple benchmark datasets are readily available, part annotations for laboratory animals require expert supervision which are both difficult and expensive to obtain. It is thus unfeasible to curate large datasets that can satiate data-hungry DCNs. Additionally, the obvious lack of visual resemblance between small laboratory animals and humans is likely to diminish the effectiveness of transfer learning methods. Unfortunately, because the body parts of small laboratory animals are far less distinctive compared to their human counterparts, the simpler HoG-based part detectors used in structured probabilistic methods such as the FMP model are likely to be inaccurate. We have confirmed experimentally that the FMP model works relatively well for unambiguous poses but fails under occlusions (see Fig.~\ref{fig:MixtureOfPartsFail} and section~\ref{evaluation}). We attribute this limitation to the local nature of the representation used to encode the appearance of parts at discrete body locations, making occlusions particularly hard to resolve. 

Herein, we describe a novel approach which extends the FMP model \cite{Yang_Ramanan_pami2013} by incorporating a measure of ``good continuity'' between adjacent parts via the incorporation of a skeleton-based shape model \cite{Trinh:Kimia:IJCV11}. However, a naive implementation would incur a significant increase in computational cost due to the need to search for ``pose candidates'' in a much higher dimensional space. We thus propose a ``cascaded'' approach in which the search space of the more complex shape-based model is restricted to the most promising pose candidates via the rapid filtering of false part configuration by a standard FMP model. We demonstrate experimentally that our approach significantly improves the accuracy of the FMP model, and outperforms a deep-learning approach \cite{wei2016cpm}.

Overall, our contributions are four-fold: {\bf (i)} We propose an extension of a representative part-based algorithm for pose estimation which incorporates a detailed representation of the body silhouette to enforce a global geometric continuity constraint between parts; To our knowledge, this is the first attempt to integrate a skeleton-based model from the shape literature \cite{Trinh:Kimia:IJCV11} with a deformable part model from the pose estimation literature \cite{Yang_Ramanan_pami2013}. {\bf (ii)} We further describe an efficient cascaded implementation for improving the efficiency of the inference process. {\bf (iii)} We thoroughly evaluate the proposed approach on a novel small laboratory animal dataset, which is shown to compare favorably with a representative deep learning baseline \cite{wei2016cpm}. {\bf (iv)} We release our software and video datasets with the hope to spur interest from the community in the application of computer vision to biomedical research.



\section{The Flexible Mixture Of Parts Model}

We first briefly describe the flexible mixture of parts (FMP) model to introduce notations (see  \cite{Yang_Ramanan_pami2013} for details). The FMP model jointly encodes the appearance and configuration of body parts and can be formalized as a graph $G = (V, E)$ such that the nodes $V$ refer to body parts and edges $E$ to their connections. When these edges only connect adjacent parts, the graph is a tree (with any node as root). 
Pose $Z$ can then defined as $Z = \{z_1, z_2, ..., z_K \}$, where $z_i = (x_i, y_i)$ describes the 2D location of body part $i$, and $K$ is the total number of parts. Appearance is defined as $T = \{\tau_1, \tau_2, ..., \tau_K \}$ where $\tau_i \in \{1, \dots P\}$ defines the identity of an appearance template $w_i^{\tau_i}$ used to encode part $i$ and is called its ``type''. Histogram of gradient (HoG) features are used as image descriptors \cite{Dalal:Triggs:CVPR05}. 

Pose estimation can then be formulated as recovering a joint configuration $Z$ and appearance $T$ that best matches an observation $I$. An appearance term is used to measure the consistency between a part appearance type $\tau_i$ and an image observation $I$ as the sum of dot-products $\sum_{i \in E} w_i^{\tau_i} \cdot \Phi(z_i, I)$ between templates $w_i^{\tau_i}$ and their image supports $\Phi(z_i, I)$. A joint configuration term is computed as the sum of the following three terms: {\bf (i)} A prior over individual parts and their types given by $\sum_{i \in E} b_i^{\tau_i}$, where $b_i^{\tau_i}$ reflects the prior probability that the part appearance template with type $\tau_i$ is assigned to part $i$.  {\bf (ii)} A prior over pairs of adjacent part types given by $\sum_{i,j \in E} b_{i,j}^{\tau_i, \tau_j}$, where $b_{i,j}^{\tau_i, \tau_j}$ reflects the dependency among types between pairs of adjacent parts.  {\bf (iii)} A prior over the relative placement of adjacent parts given by $\sum_{i,j \in E} w_{i,j}^{\tau_i, \tau_j} \cdot \Psi(z_i, z_j)$, where $\Psi(z_i, z_j)$ reflects the spatial configuration of adjacent parts and $w_{i,j}^{\tau_i, \tau_j}$ their associated weighting.

Overall, this yield the following objective function:

\begin{dmath}   
\label{eqn_yang2011}
    S (Z, \tau, I) =  \sum_{i\in E} w_i^{\tau_i} \cdot \Phi(z_i, I) + \sum_{i\in E} b_i^{\tau_i} + \sum_{i,j \in E} b_{i,j}^{\tau_i, \tau_j} + \sum_{i,j \in E} w_{i,j}^{\tau_i, \tau_j} \cdot \Psi(z_i, z_j).
\end{dmath}


\noindent Given that the configuration of body parts is modeled as a tree, the above function can be optimized efficiently via dynamic programming where a visitation schedule from children to parents is used to compute a value function for increasingly larger portions of the tree until the entire tree has been covered. This straight application of dynamic programming would typically be sufficient, except for video tracking applications because the decision made for each frame is not independent of neighboring frames. All $M$-best (rank-ordered) pose candidates  thus need to be computed, which a classic dynamic programming approach cannot provide. These $M$-best pose candidates are selected from the union of all optimal configurations backtracked from each discrete state of the selected root by considering each node as the root. Instead of simply repeating the dynamic programming $M$ times, the search for the global top $M$ candidates can then be efficiently implemented via max-product belief propagation, also referred to as two-way dynamic programming \cite{Yang_Ramanan_pami2013}. 

\begin{figure}[t!]
\begin{center}
   a)\includegraphics[height=0.44\linewidth]{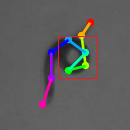}
   \includegraphics[height=0.44\linewidth]{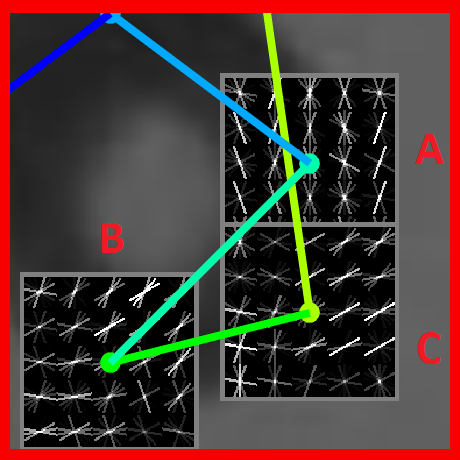}\\
   b)\includegraphics[height=0.44\linewidth]{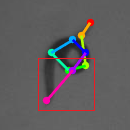}
   \includegraphics[height=0.44\linewidth]{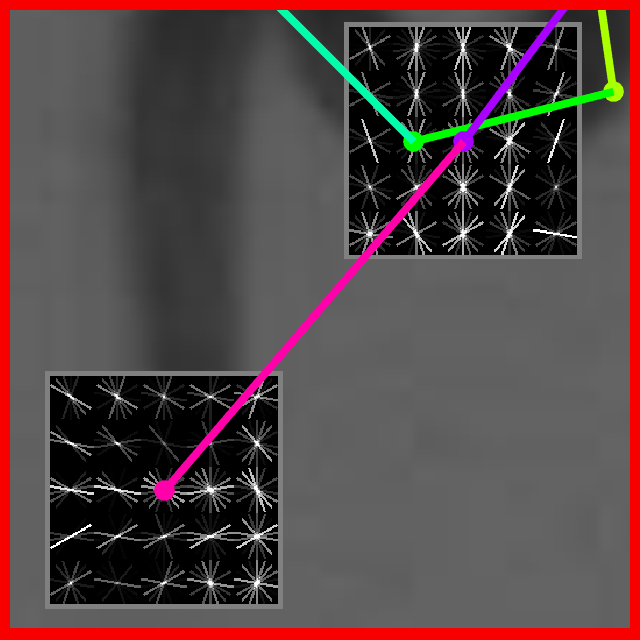}
\end{center}
\caption{Representative false detections with the FMP model. (a) An impossible pose which arises because of a violation of transitivity by the simple geometric constraints used in the FMP: (A,B) and (B, C) are placed in valid configurations but (A,C) are placed in an invalid configuration. (b) A possible but erroneous pose arises because of a lack of strong geometric constraints between adjacent parts.}
\label{fig:FMP_problems}
\end{figure}


In practice, the FMP model was shown to work well on human bodies \cite{Yang_Ramanan_pami2013}, presumably because body parts are visually distinct. However, we have found experimentally that the model fails for laboratory animals which typically lack distinct landmarks (Fig.~\ref{fig:MixtureOfPartsFail}). Also shown in Fig.~\ref{fig:FMP_problems} are representative false pose detections, which we attribute to a lack of expressiveness of the spatial constraints in the FMP model. Such spatial constraints are limited to pairs of adjacent parts, which leads to unresolvable ambiguities under occlusion. A plausible hypothesis is that enforcing an additional global constraint of geometric continuity between parts would alleviate some of these issues. This motivates our proposed extension of the FMP model to incorporate an explicit parameterization of the body silhouette such that the full shape space can be searched.




\section{Shape-consistent FMP model extension}

\begin{figure*}[!t]
\begin{center}
   a)\includegraphics[height=0.28\linewidth]{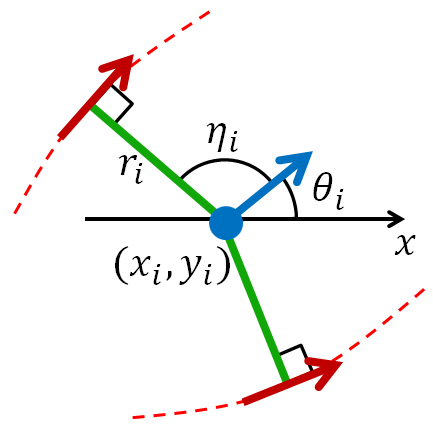}
   b)\includegraphics[height=0.28\linewidth]{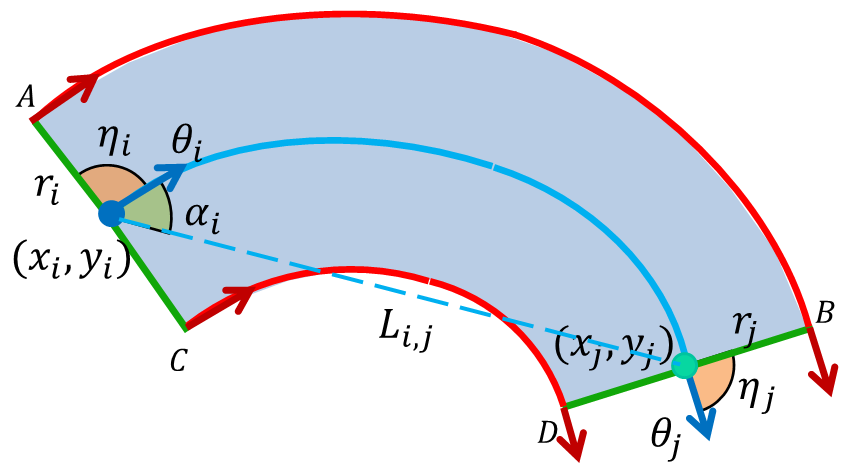}\\
   c)\includegraphics[height=0.28\linewidth]{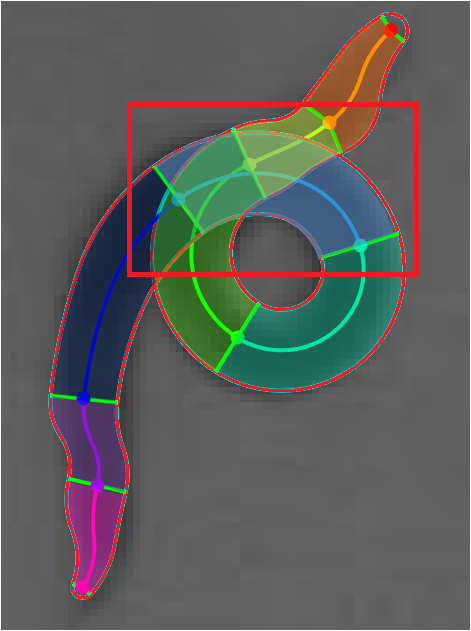}
   d)\includegraphics[height=0.28\linewidth]{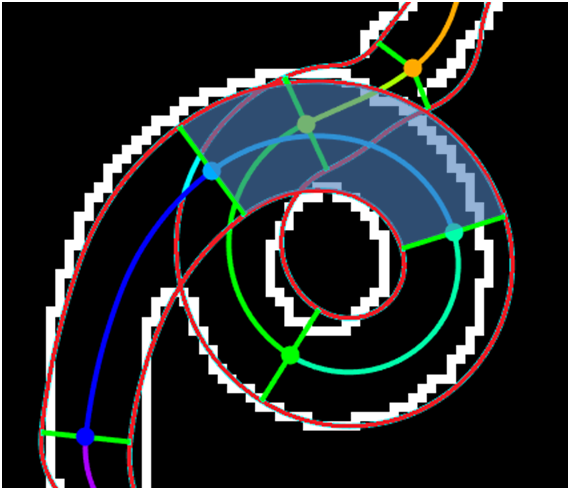}
   e)\includegraphics[height=0.28\linewidth]{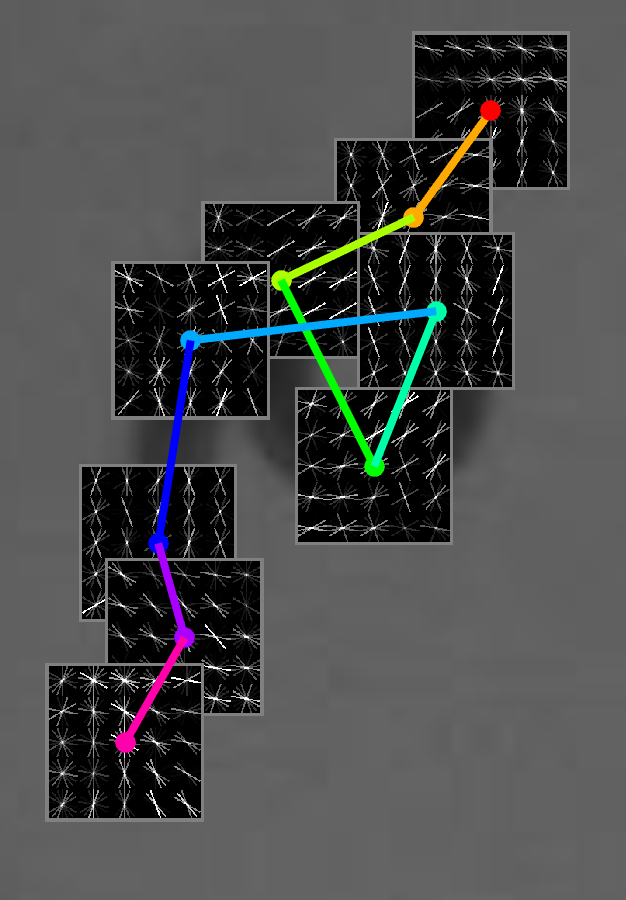}
\end{center}
   \caption{Overview of the proposed approach. (a) Model parameters include the location $(x_i,y_i)$ of each part, their orientation $\theta_i$, radius $r_i$ and flaring angle $\eta_i$. Any given combination of these 5 variables also determine the two tangents shown as red vectors which constrain a part's silhouette. (b) A shape fragment (shown as a blue region; also shown in panel c) is constraint by two adjacent parts. The blue bi-arc corresponds to the medial axis.  (c) An entire body is modeled as a collection of shape fragments (colored with different hues). (d) The consistency between measured image boundaries (white) and predicted silhouette boundaries (red) can be measured using an oriented Chamfer distance. (e) The appearance of each part is modeled as mixtures of HoG templates as in the original FMP model.}
\label{fig:shape_model}
\end{figure*}

We propose to extend the parameterization used in the original FMP model based on part locations $(x_i, y_i)$ to recover the smooth body-bounding silhouette (Fig.~\ref{fig:shape_model}c). Our parameterization uses the shape fragments formulation described in \cite{Kimia:Euler:Spiral:IJCV03} to represent the two silhouette tangents, \ie the red dotted curves in Fig.~\ref{fig:shape_model}a by their distance $r_i$ to the center of the part (constrained to be on the symmetric axis of the two tangents) and their associated angles $(\theta_i, \eta_i)$. Alternatively one can use a single angle for the part orientation defined as the average orientation of these two tangents (blue arrow in Fig.~~\ref{fig:shape_model}a, where each tangent is then represented by the angles $\pm (\eta_i \pm \pi/2)$ w.r.t. the part orientation $\psi_i$. Given a part appearance type $\tau_i$, a part $i$ can be represented by the state vector $z_i = (x_i, y_i, r_i, \theta_i, \eta_i, \tau_i)$.

The pair of silhouette boundaries between adjacent parts $z_i$ and $z_j$ are determined via bi-arc interpolation between their respective left and right tangents, \ie red contours interpolating A and B, C and D, as shown in Fig.~\ref{fig:shape_model}b  (see \cite{Kimia:Euler:Spiral:IJCV03} for details). The medial axis is represented by the blue bi-arc boundary connecting vectors $\theta_i$ and $\theta_j$. The corresponding shape fragment is shown as the blue region in Fig.~\ref{fig:shape_model}b. An entire body is defined by a graph of parts as in the original FMP model and in the skeletal shape model described in \cite{Trinh:Kimia:IJCV11}.  $C^1$ continuity of silhouette boundaries is enforced between adjacent parts. A configuration $Z = \{z_1, ..., z_K \}$ fully describes the placement of all body parts, their appearance (Fig.~\ref{fig:shape_model}e), and the global shape of the body silhouette with $C^1$ outline continuity (Fig.~\ref{fig:shape_model}c). We call this proposed model shape-consistent flexible mixture of parts model (scFMP).


\subsection*{Pose detection via scFMP}

Pose detection in the scFMP model can be formulated as finding an optimal configuration of parts $Z^*$ for a given test image $I$. From the FMP formulation, we retain the part type prior term $b_i^{\tau_i}$, the part type compatibility term $b_{i,j}^{\tau_i, \tau_j}$ and the term indicating the consistency of the appearance of an individual part with image data, \ie, $w_i^{\tau_i}\cdot \Phi(z_i, I)$. However, the availability of an explicit shape encoding in the scFMP model prompts us to modify the FMP objective function in two ways. 
First, we add a new shape term to enforce consistency between an expected silhouette shape derived from the shape fragments and an observation denoted as $\Theta(z_i, z_j, I)$ as the average oriented Chamfer distance between each proposed silhouette edge and its closest image edge, over the entire silhouette boundaries (Fig.~\ref{fig:shape_model}d). In other words, for each pixel on the red boundary of the shape fragment highlighted in blue, the closest edge point shown in white is sought and the oriented Chamfer distance is computed (the use of  a distance transform leads to a very efficient implementation), and then averaged over all proposed boundary pixels. Here, we use a classic gradient-based edge detector \cite{Kimia:Li:Guo:Tamrakar:PAMI2018} to compute image edges. 

Second, an additional shape term tries to take advantage of the additional parameters used to describe the silhouette shape to improve an image-independent ``prior'' over the relative placement of adjacent parts. Specifically, the relative placement of adjacent parts $i, j$ can be described as: (i) The distance between two parts, \ie the length of the vector $\overrightarrow{L}_{i,j}$. Since the viewing distance can vary, it is judicious to normalize the distance between parts by the radius $r_i$, leading to the term $L_{i,j}/r_i$; (ii) The angle between the vector $\overrightarrow{L}_{i,j}$ and the part orientation at node $i$, denoted as $\alpha_{i}$ (Fig.~\ref{fig:shape_model}b); (iii) A scale-invariant measure of the radius $r_j/r_i$; (iv) A bending measure between part $i$ to part $j$, \ie $\theta_i - \theta_j$;  (v) The difference in the flaring angles $\eta_i - \eta_j$. Thus the relative placement of adjacent parts can be represented as a vector $\Psi(z_i, z_j) = [L_{i,j}/r_i \;\; \alpha_{i} \;\; r_j/r_i \;\; \theta_i - \theta_j \;\; \eta_i - \eta_j]^\top$ that replaces the corresponding term in the FMP model (which uses only the first two terms in absolute coordinates). The probability of  the co-occurrence of part states $(z_i, z_j)$ can be measured as a dot-product $w_{i,j} \cdot \Psi(z_i, z_j)$. This modified measure alleviates the necessary reliance of FMP on a large amount of training data to learn the term $w_{i,j}^{\tau_i, \tau_j}$ because it no longer relies on part appearance type. 

Given a configuration of part states $Z$, the overall shape-based objective function for the scFMP model can be written as:
{\small
\begin{dmath}   
    S(Z, I) =  \sum_{i \in E} w_i^{\tau_i} \cdot \Phi(z_i, I) + \sum_{i \in E} b_i^{\tau_i} + \sum_{i,j \in E} b_{i,j}^{\tau_i, \tau_j}  + \sum_{i,j \in E} w_{i,j} \cdot \Psi(z_i, z_j) + \sum_{i,j \in E} \bar{w}_{i,j} \cdot \Theta(z_i, z_j, I).
\label{eqn_skeleton}
\end{dmath}}
\noindent The parameters $ w_i^{\tau_i}, b_i^{\tau_i}, b_{i,j}^{\tau_i,\tau_j}, w_{i,j}$ and $\bar{w}_{i,j}$ are all learned in an identical fashion to that of the traditional FMP model \cite{Yang_Ramanan_pami2013}. Because $\Theta(z_i, z_j, I)$ is a scalar, the associated weight $\bar{w}_{i,j}$ is also a scalar (unlike $w_i^{\tau_i}$ and $w_{i,j}$ which are vectors). Dynamic programming can be used for inference in order to recover the top $M$-best poses. However the increased computational cost associated with this new objective function renders the pose detection task challenging. 


\subsection*{Efficient pose detection in complex search}

One of the main challenges associated with the integration of an explicit shape model is the associated increase in computational complexity. The optimization of the objective function in pose detection depends on the size of the state space which in turn depends on the number of part states $N$, the number of appearance types $O$ and the number of parts $K$. In the traditional FMP model $N=L$, the number of image locations, while in the scFMP model $L$  is also multiplied by the number of discrete radii $r_i$ (typically 6) and the discrete number of angles $\theta_i$ and $\eta_i$ (typically 8 each). Thus, the state space increases from $L$ to $6\times8\times8\times L = 512L$, a significant increase. To make things worse, the inference gains achieved in the FMP from $\mathcal O(N^2)$ to $\mathcal O(N)$ using efficient belief propagation \cite{Felzenszwalb:Huttenlocher:IJCV05} is no longer applicable since the pairwise energy term in the scFMP formulation is now dependent on the image $I$. Thus for $L\sim 10^5$, the increase in complexity from the original FMP is $\sim (500\times10^5)^2 = 10^{13}$ which is intractable. This necessitates an efficient search algorithm. 

\begin{figure}[!t]
\begin{center}
   (a)\includegraphics[height=0.45\linewidth]{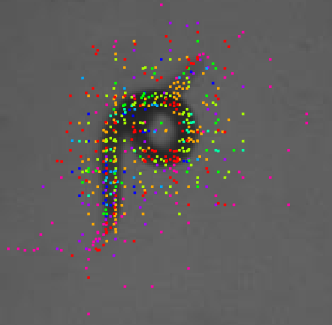}
   (b)\includegraphics[height=0.45\linewidth]{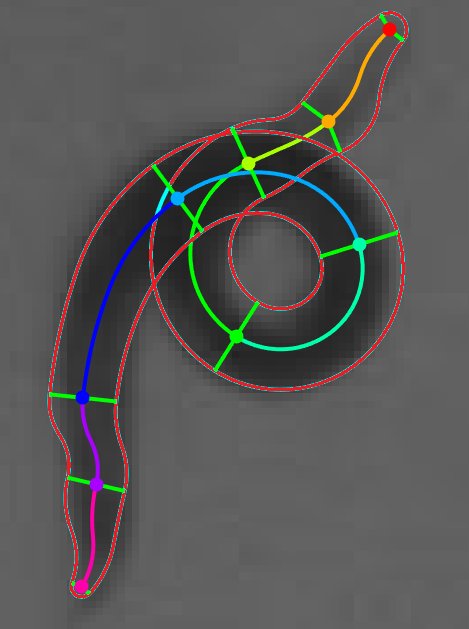}
\end{center}
   \caption{The impractically large search space for the shape-based model can be reduced by relying on part locations derived from the FMP model shown in (a), which are then augmented with additional parameters to sample the state space for our model. (b) Top-ranking pose after optimization of the shape-based model.}
\label{fig:coarse-to-fine-search}
\end{figure}

It is clear from Fig.~\ref{fig:MixtureOfPartsFail} that one of the main challenges faced by the FMP model is not that it does not find the right pose but that the right pose can sometimes be ranked low among the top $M$-best candidates. Of course, many correct solutions do not make it to the $M$-best selections, but a vast majority do; we shall see later in Fig.~\ref{fig:detection_PR}, that over 95\% of the correct part configurations are among the $M$-best candidates. Since it is often the graph connecting the parts that is incorrect but not the individual parts, we adopt an approach where the parts locations discovered by the FMP for each part serve as the state space for that part, without consideration for any connectivity between the parts (Fig.~\ref{fig:coarse-to-fine-search}). We have compared the FMP part locations against the ground-truth part locations and found that within a 4 pixel error threshold (roughly what is expected from manual annotation errors) 97\% of the ground-truth parts are present within the top $M=500$ candidates. This first stage thus defines the state space for the second stage which then optimizes the shape-based object function. 

Formally, the FMP model only provides the part locations $(x_i, y_i)$ and the next step is to compute the remaining variables $(r_i, \theta_i, \eta_i)$ that are needed for a complete characterization of a shape state. First, we observe that the distribution of radii for each part, learned from a set of training exemplars, is fairly narrow. Therefore, we simply sample from this distribution and associate a single radius $r_i$ to a pair of $(x_i, y_i)$. Second, the angle $\eta_i$ which represents the part boundary flaring from the central axis, also has a fairly narrow distribution that is learned from training exemplars. We thus sample from this distribution and associate a single value $\eta_i$ with $(x_i, y_i)$. Third, in contrast to $r_i$ and $\eta_i$, the distribution of $\theta_i$ is fairly uniform and a different strategy is needed. We observe a high degree of correlation between $\theta_i$ and the orientation of the line connecting two parts. The distribution of the angle difference between $\theta_i$ and this line's orientation, defined to be $\alpha_i$ (Fig.~\ref{fig:shape_model}b) is narrow and we therefore sample from that distribution and associate a single $\theta_i$ with $(x_i, y_i)$.

The above procedure fully constructs the state space of parts, which is restricted in size to the $M$-best locations, so that an $\mathcal O(N^2)$ algorithm can be practically implemented. We should also note that the independence of $w_{i,j}$ in our model with $\tau_i$ and $\tau_j$, in contrast to the traditional FMP model where $w_{i,j}^{\tau_i,\tau_j}$ depends on  $\tau_i$ and $\tau_j$, reduces the overall complexity from $\mathcal O(KN^2{\cal T}^2)$ to $\mathcal O(KN^2{\cal T})$, see supplementary material.



\subsection*{Shape-based pose tracking}

Single frames are often ambiguous and the single-best part configuration is not always correct. As a result, one needs to consider not just the single-best but the top $M$-best candidate poses $Z_t$ from each frame $I_t$. This typically leads to a very large number of possible paths for an entire video sequence ${\cal Z} = \{Z_1, Z_2, \cdots, Z_T\}$. Tracking aims at recovering an optimal sequence of poses ${\cal Z^*}$ by enforcing temporal continuity between frames while simultaneously optimizing the selection of $M$-best poses for each frame individually. Specifically, an $M$-best selection is rated according to $S_{frame}(Z_t, I_t)$ as defined in Eq.~\ref{eqn_skeleton}, while temporal smoothness is computed as $S_{pairwise}(Z_{t-1}, Z_t)$ which can be simply computed as the sum of squared distance between corresponding parts at adjacent time stamps. These two terms give rise to an objective function over sequence ${\cal Z}$ in the form:

\begin{dmath}
S({\cal Z}) = \sum_{t=1}^T S_{frame} (Z_t, I_t) +  \gamma \sum_{t=2}^{T-1} S_{pairwise} (Z_{t-1}, Z_t),
\label{eqn_temp_dp}
\end{dmath}
where $\gamma$ balances the two energies. This objective function can be maximized using dynamic programming. Here, we introduced a specific interpolated pose representation specifically for nematodes which is described in the supplementary material.


\section{Evaluation}
\label{evaluation}

Next, we evaluate the accuracy of the proposed scFMP model using both a frame-based accuracy measure (where individual frames are treated independently) and a video tracking accuracy measure (where individual frames are embedded in a tracking sequence).  One standard criterion for pose detection is the percentage of correct keypoint (PCK) introduced in \cite{Yang_Ramanan_pami2013}. A candidate keypoint is correct if it falls within $\beta \cdot \max(h,w)$ pixels of the ground-truth location, where $h$ and $w$ are the height and width of the smallest rectangular window that includes all the keypoints of the ground-truth pose. The maximum PCK is reported when multiple pose candidates are generated and compared to the ground-truth.  We argue that a proper evaluation should not only take into account the maximum score across all candidate poses, but also their overall accuracy because when candidate poses are passed to the next tracking stage, the higher the overall quality of the candidate poses is, the less ambiguity is left for the tracker to resolve. Based on this argument, we propose a criterion to evaluate pose estimation via a mean PCK -- max PCK curve, which is produced by varying the number $M$ of candidate poses produced by the model.



\begin{figure*}[t!]
\begin{center}
   \includegraphics[width=0.15\linewidth]{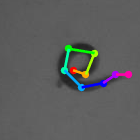}
   \includegraphics[width=0.15\linewidth]{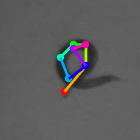}
   \includegraphics[width=0.15\linewidth]{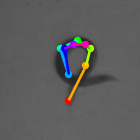}
   \includegraphics[width=0.15\linewidth]{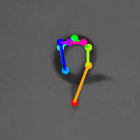}
   \includegraphics[width=0.15\linewidth]{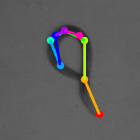}
   \includegraphics[width=0.15\linewidth]{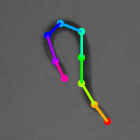}\\
   \includegraphics[width=0.15\linewidth]{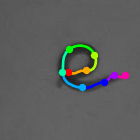}
   \includegraphics[width=0.15\linewidth]{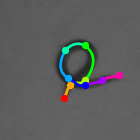}
   \includegraphics[width=0.15\linewidth]{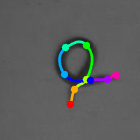}
   \includegraphics[width=0.15\linewidth]{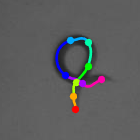}
   \includegraphics[width=0.15\linewidth]{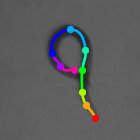}
   \includegraphics[width=0.15\linewidth]{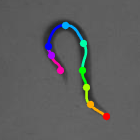}
\end{center}
   \caption{Optimal sequence of poses derived from the original FMP (top) the proposed scFMP extension (bottom).}\label{fig:visual_compare_pose_seq:worm}
\end{figure*}

\subsection{Experiments on nematodes datasets}

\begin{figure}[ht!]
\begin{center}
   a) \includegraphics[width=0.95\linewidth]{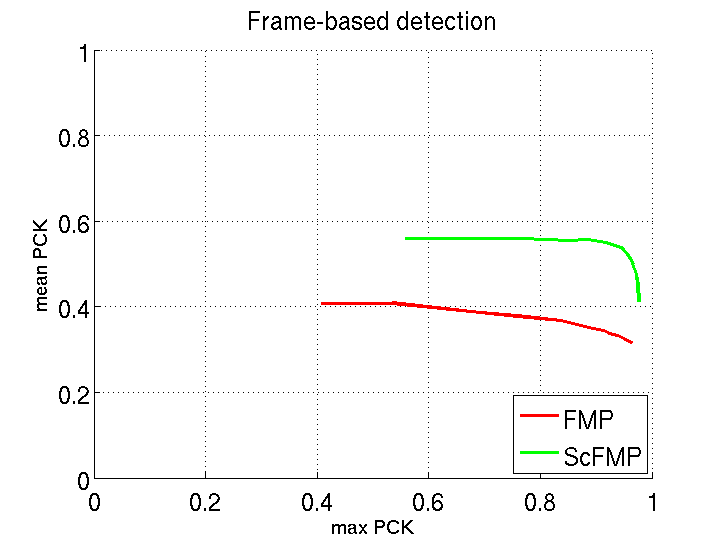}\\
   b) \includegraphics[width=0.95\linewidth]{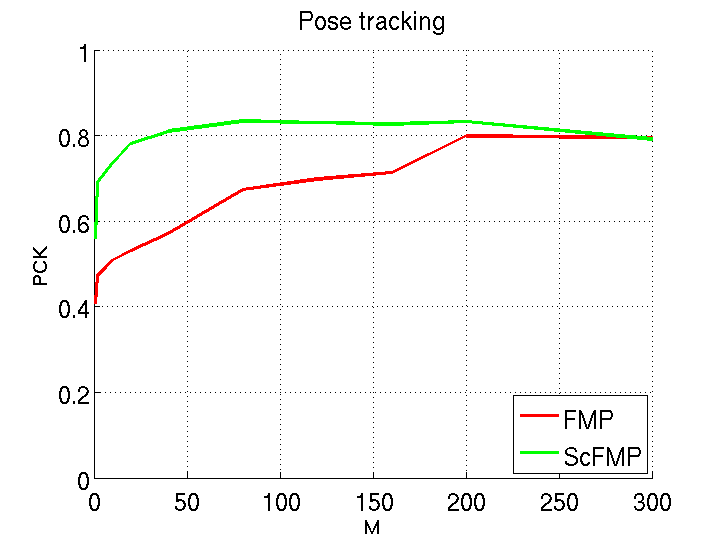}
\end{center}
   \caption{Evaluation on the nematodes dataset. (a) Frame-based accuracy measured as mean PCK vs. max PCK (see text for details) generated by varying the number of $M$-best poses returned by the algorithm. (b) Sequence-based tracking accuracy measured as mean PCK over sequence vs. number of extracted poses $M$.}\vspace{3mm}
\label{fig:detection_PR}
\end{figure}

We collected ten representative video sequences of nematodes (each corresponding to a different animal) with a total length of 1,200 frames (30 fps). Ground-truth poses were manually annotated every 10 frames by marking the location of body parts from head to tail (9 discrete locations). Five videos were selected as the training set and the other five videos as the test set (Fig.~\ref{fig:detection_PR}a). The scFMP extension significantly outperforms the original FMP model. We also evaluate the overall accuracy of the tracking pipeline by measuring the average PCK over entire video sequences as a function of the number $M$ of pose sequences returned by the FMP and the scFMP models. As shown in Fig.~\ref{fig:detection_PR}b, the  proposed scFMP extension again consistently outperforms the original FMP model and, for both models, our tracker is shown to significantly improve accuracy. Qualitative comparison between optimal pose sequences derived from the FMP and the scFMP models are shown in Fig.~\ref{fig:visual_compare_pose_seq:worm}. The visual comparison shows that the proposed approach is far more robust in handling self-occlusion, and is capable of recovering the continuous deformation of a body under occlusion.

\begin{figure*}[!t]
\begin{center}
   \includegraphics[width=0.19\linewidth]{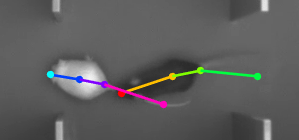}
   \includegraphics[width=0.19\linewidth]{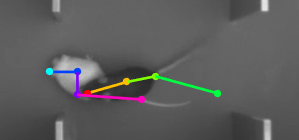}
   \includegraphics[width=0.19\linewidth]{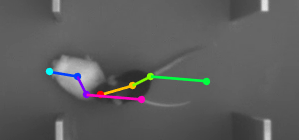}
   \includegraphics[width=0.19\linewidth]{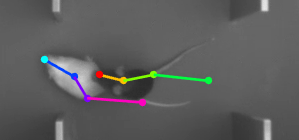}
   \includegraphics[width=0.19\linewidth]{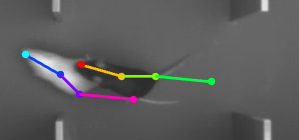}\\
   \includegraphics[width=0.19\linewidth]{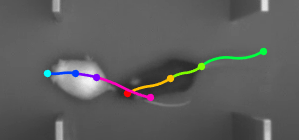}
   \includegraphics[width=0.19\linewidth]{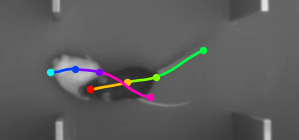}
   \includegraphics[width=0.19\linewidth]{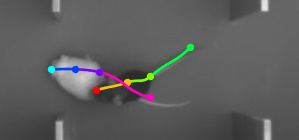}
   \includegraphics[width=0.19\linewidth]{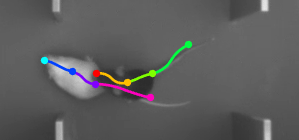}
   \includegraphics[width=0.19\linewidth]{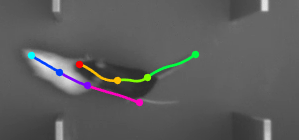}\\
\end{center}
   \caption{(Top) Optimal pose sequence derived using the original FMP. (Bottom) Optimal pose sequence from the proposed scFMP extension.}
\label{fig:visual_compare_pose_seq:mice}
\end{figure*}
\subsection*{Comparison with the Convolutional Pose Machines}

The Convolutional Pose Machine (CPM) is a representative deep learning-based approach that performs pose estimation from a set of part-based detections \cite{wei2016cpm}.  We used publicly available code\footnote{\url{https://github.com/timctho/convolutional-pose-machines-tensorflow.git}} and fine-tuned the model pre-trained on hand pose for our dataset. CPMs were trained on object-centered bounding boxes and the network was then turned into a fully-convolutional network. $1\times1$ max-pooling was performed over all overlapping output confidence maps to get global confidence scores. To serve as a positive control, we initially placed a $10\times10$ colored patch, color-coded by landmark ID, on each anatomical keypoint in both train and test images. This yielded a meanPCK score of 0.997, validating the model and implementation. A comparison with our system on actual nematode images is shown in Table~\ref{tab:CPM}. 

\begin{table}[h!]
    \centering
    \begin{tabular}{|c|c|c|c|}
        \hline
        & CPM \cite{wei2016cpm} & ours ($M=1$) & ours ($M=80$) \\
        \hline
        PCK & 0.59 & 0.56 & 0.83 \\
        \hline
    \end{tabular}
    \caption{Accuracy of the scFMP model vs. a CPM baseline.}
    \label{tab:CPM}\vspace{-3mm}
\end{table}

Because CPMs yield only a single output candidate pose, we report evaluation scores for scFMP with $M=1$ (for direct comparison) as well as our complete system ($M=80$). Additionally, we trained a model on edge magnitudes concatenated with pixel intensities, to explicitly test the importance of partial shape information. Though this did not help performance (meanPCK = $0.50$), we hypothesize that this might be due to a lack of sufficient global constraints given that boundaries of the arena as well as background artifacts serve as strong deterrents.

\subsection{Experiments on rodents datasets}

\begin{figure}[t!]
\begin{center}\vspace{10mm}
   a) 
   \includegraphics[width=0.95\linewidth]{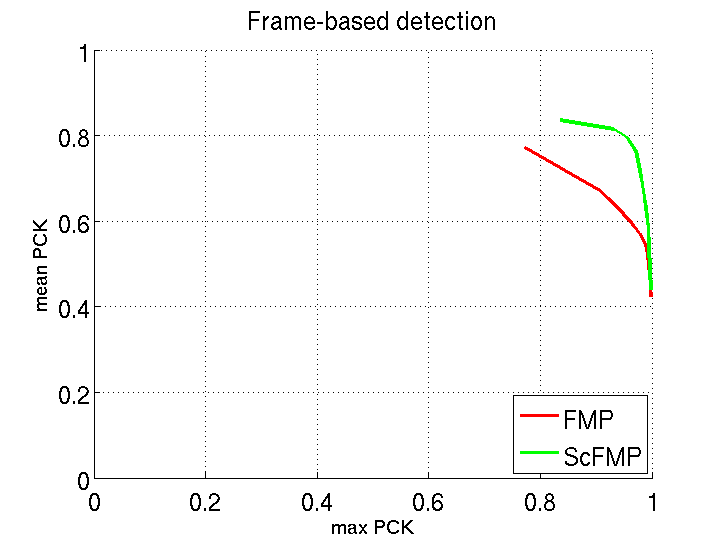}\\
   b) 
   \includegraphics[width=0.95\linewidth]{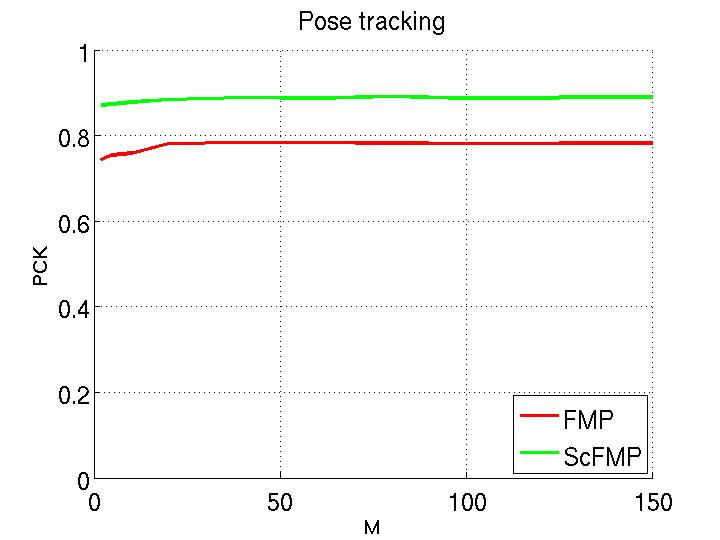}
\end{center}
   \caption{Evaluation on the rodents dataset. (a) Frame-based accuracy measured as mean PCK vs. max PCK (see text for details) generated by varying the number of $M$-best poses returned by the algorithm. (b) Sequence-based tracking accuracy measured as mean PCK over sequence vs. number of extracted poses $M$.}\vspace{15mm}
\label{fig:detection_PR:mice}
\end{figure}

As an additional validation of the system, we also collected ten video sequences containing rodents. Each video recording also included 1,000 frames (30 fps) with ground-truth poses manually annotated every 10 frames at the following keypoints: head, body, hip, and tail tip. Five videos were taken as the training set and the other five as the testing set. We performed the same evaluation as done for nematode videos based on both the accuracy of frame-based pose detection and the accuracy for the optimal sequence of poses from a video. The experimental results shown in Fig.~\ref{fig:detection_PR:mice} confirm the superiority of the extended scFMP over the original FMP. Qualitative comparison between optimal pose sequences derived from the FMP and the scFMP models are shown in Fig.~\ref{fig:visual_compare_pose_seq:mice}. The proposed approach is more robust to occlusions and more accurate at localizing thin body parts, \eg the tail. 


\section{Conclusion}
 
We have proposed an extension of the FMP model  for estimating body pose to incorporate an explicit shape model. A ``cascade'' formulation is described to make the approach more efficient. We experimentally showed that the proposed approach yields a very significant improvement over the original FMP model as well as a representative deep convolution neural network, especially under occlusion. 
 
\section*{Acknowledgments}
This work was supported by the Robert J. and Nancy D. Carney Fund for Scientific Innovation to TS. Additional funding was provided by the Human Frontier Science Program (RGP0006/2015), DARPA young faculty award [grant number N66001-14-1-4037] and NSF early career award [grant number IIS-1252951] to TS. TS serves on the scientific advisory board of Vium, Inc.
{\small
\bibliographystyle{ieee}
\bibliography{Paperpile,proceedings}
}

\clearpage

{\bf\Large\centering Supplementary Material}
\vspace{5mm}

In this supplementary material, we first include a detailed discussion of the complexity of the proposed scFMP approach. We provide additional details regarding the implementation of the scFMP and further motivate the introduction of a shape-based temporal consistency measure, which is particularly helpful for tracking highly deformable body movements.






\section*{Computational complexity in pose detection via scFMP}
\label{sec:complexity:scFMP}

Recall that the objective function for the proposed shape-consistent mixture of parts (scFMP) model can be written as:

{\small
\begin{dmath}   
    S(Z, I) =  \sum_{i \in E} w_i^{\tau_i} \cdot \Phi(z_i, I) + \sum_{i \in E} b_i^{\tau_i} + \sum_{i,j \in E} b_{i,j}^{\tau_i, \tau_j}  + \sum_{i,j \in E} w_{i,j} \cdot \Psi(z_i, z_j) + \sum_{i,j \in E} \bar{w}_{i,j} \cdot \Theta(z_i, z_j, I).
\label{eqn_skeleton}
\end{dmath}}


Observe that the descriptor $\Phi(z_i, I)$ is independent of the part type $\tau_i$, the descriptors $\Psi(z_i, z_j) and \Theta(z_i, z_j, I)$ are independent of part types $\tau_i, \tau_j$. By writing $s_i = (x_i, y_i, r_i, \theta_i, \eta_i)$ and the state of each part $i$ as $z_i = (s_i, \tau_i)$, the descriptors can thus be re-written as $\Phi(s_i, I) = \Phi(z_i, I)$,  $\Psi(s_i, s_j) = \Psi(z_i, z_j)$ and $\Theta(s_i, s_j, I) = \Theta(z_i, z_j, I)$. The objective function
in turn can be re-written as:
{\small
\begin{align}\label{eqn_skeleton_short}
    S(Z, I) & =  \sum_{i \in E} {\cal \Phi}(z_i, I) + \sum_{i,j \in E} {\cal \Psi}(z_i, z_j, I),\\
    \mbox{where}\; &{\cal \Phi}(z_i, I) = b_i^{\tau_i} + w_i^{\tau_i} \cdot \Phi(s_i, I) \nonumber\\
            &{\cal \Psi}(z_i, z_j, I) = b_{i,j}^{\tau_i, \tau_j} + w_{i,j} \cdot \Psi(s_i, s_j) + \bar{w}_{i,j} \cdot \Theta(s_i, s_j, I)\nonumber.
\end{align}}

The optimization of the objective function can be efficiently implemented via dynamic programming where all the parts are iteratively visited following a schedule starting from the leaves and moving upstream to the root part. For each particular part $i$, the message passed to its parent part $j$ is computed as:

\begin{align}\label{eqn_message}
    \mbox{score}_i(z_i) &= {\cal \Phi}_i(z_i, I) + \sum_{k\in kids(i)} m_k(z_i)\\
    \mbox{where} \;\; m_k(z_i) &= \max_{z_k}[\mbox{score}_k(z_k) + {\cal \Psi}_{k,i}(z_k, z_i)].
\end{align}

The computational complexity of the dynamic programming can be derived from an analysis of the complexity of each iteration in computing Eq.~\ref{eqn_message}, such that $m_k(z_i)$ can be expanded as:

\vspace{5mm}

\begin{dmath}
    m_k(s_i, \tau_i) = \max_{\tau_k} \{b_{k,i}^{\tau_k, \tau_i} + 
                       \max_{s_k} [score_k(s_k, \tau_k)  +  w_{k,i} \cdot \Psi(s_k, s_i)
                         + \bar{w}_{k,i} \cdot \Theta(s_k, s_i, I)]\}.
\end{dmath}

Given $N$ the number of states per part and $\cal T$  the number of appearance types per part, the complexity in computing $m_k(s_i, \tau_i)$ is $O(N^2{\cal T})$, and in turn the complexity in computing $\mbox{score}_i(z_i)$ is $O(N^2 {\cal T} + N {\cal T})=O(N^2{\cal T})$. In contrast to the traditional FMP, where $w_{i,j}^{\tau_i,\tau_j}$ depends on  $\tau_i$ and $\tau_j$, our $w_{i,j}$ is independent of $\tau_i$ and $\tau_j$, which reduces the complexity in computing $\mbox{score}_i(z_i)$ from $O(N^2{\cal T}^2)$ to $O(N^2{\cal T})$. Considering $K$ parts in total, the overall complexity in optimizing the proposed shape-based objective function is $O(KN^2{\cal T})$.


\section*{Learning of the scFMP model}
\label{sec:learning}

In order to use the shape-based objective function (Eq.~\ref{eqn_skeleton}) for pose detection, the model parameters, namely, $b_i^{\tau_i}, b_{i,j}^{\tau_i, \tau_j}, w_i^{\tau_i}, w_{i,j}, \bar{w}_{i,j}$ have to be learned appropriately such that the score of the objective function has high value for the right poses. Our learning scheme directly follows the one proposed in \cite{Yang_Ramanan_pami2013}. Our training data includes positive poses annotated for each animal and negative examples generated by running a detector within images without any target object. A structured prediction objective function can thus be formulated as done in \cite{Felzenszwalb:etal:PAMI09,Yang_Ramanan_pami2013}. Let $ \beta $ denote all the model parameters, $b_i^{\tau_i}, b_{i,j}^{\tau_i, \tau_j}, w_i^{\tau_i}, w_{i,j}, \bar{w}_{i,j}$,  and let $\Gamma(Z,I)$ denote all the energy terms, namely $\Phi(s_i, I), \Psi(s_i, s_j), \Theta(s_i, s_j, I)$, scoring function~\ref{eqn_skeleton} can be written as $ E(Z,I) = \beta \cdot \Gamma(Z,I) $. The model parameters are learned following the formulation of structural SVM:

\begin{eqnarray}
    arg \min_{w, \xi_i \geq 0} \frac{1}{2} \beta \cdot \beta + \sum_n \xi_n,\\
    s.t. \quad \forall n \in pos \quad \beta \cdot \Gamma(Z_n,I_n) \geq 1 - \xi_n \nonumber\\
    \quad \forall n \in neg, \forall z \quad  \beta \cdot \Gamma(Z,I_n) \leq -1 + \xi_n, \nonumber    
\end{eqnarray}
where the constraints state that positive examples should be scored higher than 1, while negative examples should be scored lower than -1. This objective function penalizes violations of these constraints using slack variables $\xi_n$. The above optimization problem is in turn formulated as a quadratic program (QP) problem that is solved via a dual coordinate-descent solver introduced in \cite{Yang_Ramanan_pami2013}.


\section*{Shape-based temporal smoothness term for nematodes}
\label{sec:shape:tracking}

\begin{figure}[!ht]
\begin{center}
   \quad\includegraphics[width=0.2\linewidth]{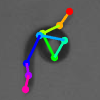}
   \quad\includegraphics[width=0.2\linewidth]{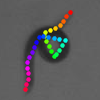}
   \quad\includegraphics[width=0.2\linewidth]{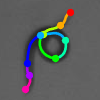}
   \quad\includegraphics[width=0.2\linewidth]{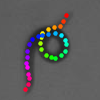}\\
   a)\,\includegraphics[width=0.2\linewidth]{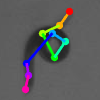}
   b)\,\includegraphics[width=0.2\linewidth]{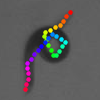}
   c)\,\includegraphics[width=0.2\linewidth]{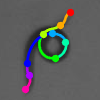}
   d)\,\includegraphics[width=0.2\linewidth]{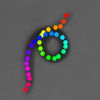}
\end{center}
   \caption{(a) The default pairwise energy term penalizes the distance between parts (top row) and its subsequent pose (bottom row). This model is sensitive to the distance between parts since variations over large segments of the shape are penalized in the same way as variations over small segments. (b) Resampling the graph and penalizing variations over all points alleviates this problem. (c) Representing shape by a line between two parts is not as accurate as a curved interpolation. (d) Combining the two ideas described in b and c. }
\label{fig:temporal_consistency}
\end{figure}

Directly applying the default temporal smoothness term, $S_{pairwise}(Z_{t-1}, Z_t)$, to the nematodes faces two challenges: (i) the distribution of the body parts over the object is not necessarily uniform: some parts are close while others are far. The above energy unfairly under emphasizes the role of parts with large inter-part distances. This is illustrated in Fig.~\ref{fig:temporal_consistency}a where detections from a pair of consecutive frames is shown, one on top of the other. Observe how the yellow segment is small while the blue segment is twice its size. Since each segment represents an object proportional to its length, it makes sense to re-sample the graph so that all nodes are roughly the same distance from their neighbors, Fig.~\ref{fig:temporal_consistency}b, which demonstrably improves results; (ii) When the object articulates, the graph links do not necessarily fit the body well: the straight line localization of Fig.~\ref{fig:temporal_consistency}a is far from the underlying object, this in turn causes unexpected penalty as well. Thus, we use bi-arc interpolation to reach a curved representation of the object that closely follow the natural shape of the body, as in Fig.~\ref{fig:temporal_consistency}c.


\end{document}